\documentclass[runningheads]{llncs}
\usepackage[T1]{fontenc}
\usepackage{graphicx}
\usepackage{amsfonts}  % lightweight mathbb
\begin{document}
\title{LEARN: A Story-Driven Layout-to-Image Generation Framework for STEM Instruction}
\titlerunning{LEARN: A Story-Driven Layout-to-Image Generation Framework}
\author{Maoquan Zhang\inst{1,2}\orcidID{0009-0001-4356-3503} \and
Bisser Raytchev\inst{1}\orcidID{0000-0002-2146-415X} \and
Xiujuan Sun\inst{2}\orcidID{0009-0009-9151-1052}}

\authorrunning{Maoquan Zhang et al.}
\institute{
Graduate School of Advanced Science and Engineering, Hiroshima University, Hiroshima 739-8511, Japan\\
\email{zhang-maoquan@hiroshima-u.ac.jp}, \email{bisser@hiroshima-u.ac.jp}
\and 
Department of Computer Science, Weifang University of Science and Technology, Shouguang 262700, China\\
\email{xjs1981524@wfust.edu.cn}
}

\maketitle              % typeset the header of the contribution
\begin{abstract}
LEARN is a layout-aware diffusion framework designed to generate pedagogically aligned illustrations for STEM education. It leverages a curated BookCover dataset that provides narrative layouts and structured visual cues, enabling the model to depict abstract and sequential scientific concepts with strong semantic alignment. Through layout-conditioned generation, contrastive visual–semantic training, and prompt modulation, LEARN produces coherent visual sequences that support mid-to-high-level reasoning in line with Bloom’s taxonomy while reducing extraneous cognitive load as emphasized by Cognitive Load Theory. By fostering spatially organized and story-driven narratives, the framework counters fragmented attention often induced by short-form media and promotes sustained conceptual focus. Beyond static diagrams, LEARN demonstrates potential for integration with multimodal systems and curriculum-linked knowledge graphs to create adaptive, exploratory educational content. As the first generative approach to unify layout-based storytelling, semantic structure learning, and cognitive scaffolding, LEARN represents a novel direction for generative AI in education. The code and dataset will be released to facilitate future research and practical deployment.

\end{abstract}
\section{Introduction}

Effective STEM education requires more than delivering factual content. It depends on guiding learners across cognitive levels—from memorization to synthesis and application—as framed by Bloom's Taxonomy of Learning Domains~\cite{bloomrevision2002}. Visual materials such as conceptual layouts and analogical illustrations are essential for externalizing abstract ideas and sustaining cognitive engagement. However, most current instructional visuals lack narrative continuity, spatial precision, and adaptability to different learning trajectories.

Cognitive Load Theory (CLT)~\cite{paas2003cognitive,sweller1988cognitive} underscores that visual input should reduce cognitive friction, align with learner processing, and provide scaffolds for building mental models. In STEM domains, where relationships such as causality, symmetry, and constraint are central, the absence of expressive and pedagogically aligned imagery can hinder comprehension. Generative visual tools that adapt to both content and cognitive needs represent a promising but underexplored avenue.

Recent advances in text-to-image generation~\cite{2022hierarchical_t2i,saharia2022photorealistic} enable open-domain synthesis, yet these models are structurally shallow and unsuitable for educational use. Layout-to-image (L2I) approaches~\cite{li2023gligen,zheng2023layoutdiffusion,cheng2024rethinking} provide stronger spatial control but often lack semantic depth and rarely encode instructional scaffolding. For example, SDXL-RC~\cite{cheng2024rethinking} advances layout-rich generation but does not incorporate curriculum logic or cognitive progression, which are critical for effective STEM illustrations.

To address this gap, LEARN (\textbf{L}ayout-\textbf{E}nabled \textbf{A}utomatic \textbf{R}endering of \textbf{N}arratives) is introduced as a framework for generating educational illustrations with explicit layout awareness and pedagogical alignment. At its core is a curated BookCover dataset designed to represent abstract and multi-step scientific concepts in visually structured forms. The initial data were adapted from the public Book Covers dataset on Kaggle~\cite{bookcovers_kaggle} and extended with annotations to support instructional goals. For each image, bounding boxes were generated using CLIPSeg~\cite{2022clipseg} and SAM~\cite{ravi2024sam2segmentimages}, and high-level semantic descriptions of object relationships were added via GPT-4o. These enriched annotations allow the model to learn spatial composition alongside the pedagogical rationale behind object arrangements. Fine-tuning on this dataset enables LEARN to render abstract STEM concepts as visually coherent, cognitively supportive sequences that align with Bloom’s taxonomy and reduce extraneous cognitive effort as emphasized by CLT.

The framework combines layout-conditioned diffusion, prompt modulation, and semantic alignment to produce story-driven illustrations with compositional precision and instructional clarity. Experiments on STEM-specific prompts demonstrate that LEARN can generate scalable, narrative-rich visual content tailored to educational needs. This work is, to our knowledge, the first layout-aware image generation framework explicitly integrating educational theory, cognitive scaffolding, and dataset design into a unified generative AI system for STEM instruction.

%-----------------20250505----
%-----------------20250506----
\section{Related Work}
\subsubsection{Educational Visual Representations}
The role of visualizations in supporting learning and comprehension has been extensively studied across cognitive psychology and educational research. Dual coding theory~\cite{paivio2014dual} suggests that representing information in both verbal and visual forms enhances recall and conceptual integration. In STEM education, diagrammatic representations, schematic layouts, and analogical illustrations are known to reduce extraneous cognitive load~\cite{sweller1988cognitive,mayer2005cambridge}, promote inferential reasoning~\cite{ainsworth2006deft}, and facilitate transfer of learning~\cite{goldstone2003transfer}. However, such visual aids are typically handcrafted, static, and fail to scale with diverse learner needs. Our work builds on this tradition by introducing a generative framework that learns from rich image-text pairs and automatically composes layouts tailored to educational narratives.
\subsubsection{Layout-to-Image (L2I) Generation}
Layout-conditioned generation has emerged as a promising paradigm for controllable image synthesis, where semantic layouts (defined as object labels with bounding boxes) serve as structural priors. Earlier methods used GANs~\cite{hong2018inferring,li2019object}, while recent models rely on diffusion-based architectures~\cite{cheng2024rethinking,zheng2023layoutdiffusion} to enhance visual fidelity. However, most L2I models assume the availability of manually specified layouts, and are primarily evaluated on datasets like COCO-Stuff or Visual Genome, where the layouts serve aesthetic or scene-structuring purposes rather than educational semantics. In contrast, we propose to infer layouts automatically from concept descriptions and use them as cognitive scaffolds in instructional visual generation.
\subsubsection{Image-Text Alignment and Multimodal Reasoning}
Large-scale vision-language models, such as CLIP~\cite{radford2021clip}, BLIP~\cite{li2022blip}, and Flamingo~\cite{alayrac2022flamingovisuallanguagemodel}, have demonstrated impressive alignment between text and image modalities, enabling flexible conditioning of generative models. These alignments, however, are typically unstructured and optimized for retrieval or open-domain generation. More structured grounding approaches, such as GLIGEN~\cite{li2023gligen} and layoutDiffusion~\cite{zheng2023layoutdiffusion}, incorporate textual and positional prompts simultaneously, but are not designed to disentangle pedagogically meaningful object relations. Our model extends this line of work by grounding educational concepts not just in objects or attributes, but in layout configurations that convey meaning through spatial hierarchy, juxtaposition, and narrative cohesion.
\subsubsection{Narrative and Knowledge-Driven Generation}
Recent efforts have begun to explore concept- or knowledge-graph-driven generation~\cite{peng2019cm,text2scene2019}, where structured knowledge guides the content of generated scenes. However, few works have addressed educational settings, where the alignment between conceptual structure, layout coherence, and learner cognition is critical. Book covers, with their rich interplays between visual metaphor and thematic storytelling, offer an untapped resource for learning such compositional mappings. Our system uses this resource to build layout-aware mappings from instructional narratives to visual representations that support cognitive progression across Bloom's taxonomy~\cite{bloomrevision2002}.

\section{Method}
\begin{figure}[t]
\centering
\includegraphics[width=\textwidth, trim=0cm 0cm 0cm 0cm, clip]{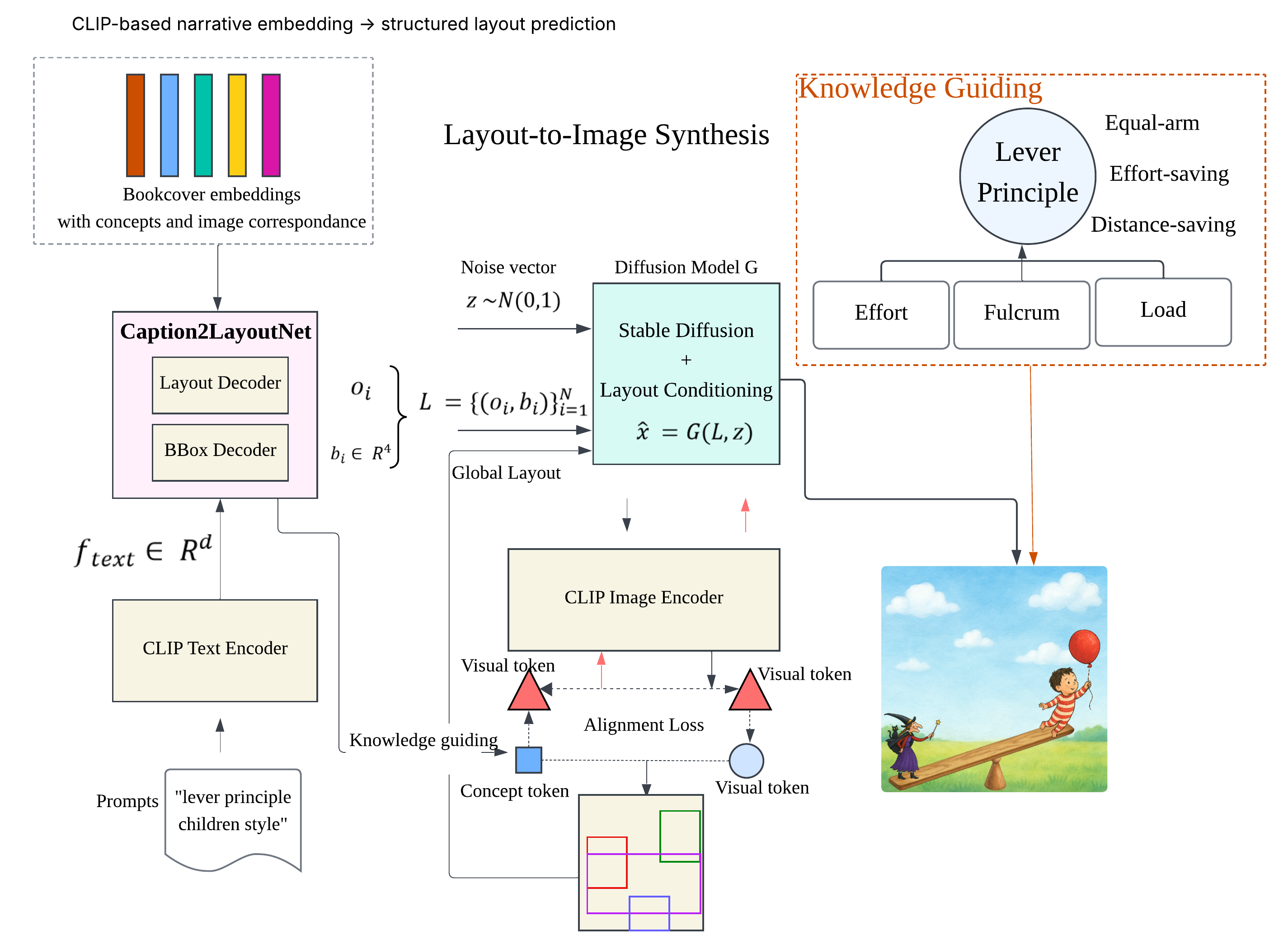}
\caption{Overview of the LEARN framework. A STEM concept prompt (e.g., “lever principle”) is encoded using a CLIP text encoder to obtain a semantic embedding. Caption2LayoutNet generates a structured layout comprising object tokens and bounding boxes. This layout, together with a sampled noise vector, conditions a diffusion model to synthesize a concept-relevant image. The generated image is then aligned with the original prompt via a self-supervised contrastive loss. Knowledge-guided layout priors are learned from a curated BookCover dataset, linking visual structure to conceptual content.}
\label{fig1}
\end{figure}

Here we present the core components of our LEARN framework for concept-grounded visual generation in STEM education, which consists of three modules: (1) \textbf{Narrative Encoding and Layout Generation}, (2) \textbf{Layout-to-Image Synthesis}, and (3) \textbf{Knowledge-driven Iterative Visualization}. As illustrated in Fig.~\ref{fig1}, our pipeline transforms a STEM concept into a structured layout via \textbf{Caption2LayoutNet}, synthesizes a corresponding image through layout-conditioned diffusion, and enforces semantic consistency using CLIP-based alignment losses guided by BookCover-derived visual–concept associations.

\subsection{Narrative Encoding and Layout Generation}
Given a textual description of an abstract STEM concept (e.g., "Lever Principle"), our goal is to generate a spatial layout that organizes key entities and relations for visual rendering.  Inspired by grounded generation techniques, the \textbf{Caption2LayoutNet} module is introduced. It consists of a transformer-based decoder guided by contrastive alignment to BookCover semantics, and is conditioned on a pre-trained CLIP-based text encoder $f_{{text}}(\cdot)$ and a learnable layout decoder $f_{{layout}}(\cdot)$:
\begin{equation}
L = f_{{layout}}(f_{{text}}(c))
\end{equation}
where $c$ denotes the input concept sentence, and $L = {(o_i, b_i)}_{i=1}^N$ represents the set of predicted object labels $o_i$ and corresponding bounding boxes $b_i$. Each $b_i \in \mathbb{R}^4$ encodes the position and size of the object in the image canvas, with $b_i = (x_i, y_i, w_i, h_i)$ indicating the normalized top-left coordinates and size dimensions.

To enable layout-aware generation, each layout element—defined by label $o_i$ and bounding box $b_i$—is encoded into a layout embedding $l_i \in \mathbb{R}^d$ using CLIP-based semantic encoding and positional embedding:
\begin{equation}
l_i = f_{{label}}(o_i) + f_{{pos}}(b_i)
\end{equation}
Note that $o_i$ denotes the symbolic object label (e.g., "ball", "magnet"), and $l_i$ refers to its layout embedding in $\mathbb{R}^d$ space. To avoid confusion, we reserve $o_i$ for labels and $l_i$ for the corresponding embeddings throughout the paper.
To ensure pedagogical plausibility, a \textbf{self-supervised alignment loss} is applied to encourage consistency between the predicted layout embedding $l_i$ and the visual semantics of real BookCover images. Specifically, region-level visual embeddings $v_i$ are extracted from BookCover images using a frozen CLIP image encoder, and matched with predicted layout embeddings $l_i$. The alignment loss is defined as:
\begin{equation}
\mathcal{L}_{\mathrm{align}} = -\frac{1}{N} \sum_{i=1}^{N} \log \frac{\exp(\mathrm{sim}(l_i, v_i)/\tau)}{\sum_{j=1}^{N} \exp(\mathrm{sim}(l_i, v_j)/\tau)}
\end{equation}
where $\mathrm{sim}(\cdot, \cdot)$ denotes similarity and $\tau$ is a temperature hyperparameter. $N$ is the number of predicted layout components within a single image. This contrastive formulation ensures that each predicted layout embedding $l_i$ aligns closely with its corresponding visual region $v_i$ while being distinguishable from others.

\begin{figure}[t]
\centering
\includegraphics[width=\textwidth, trim=0cm 0.1cm 0cm 0cm, clip]{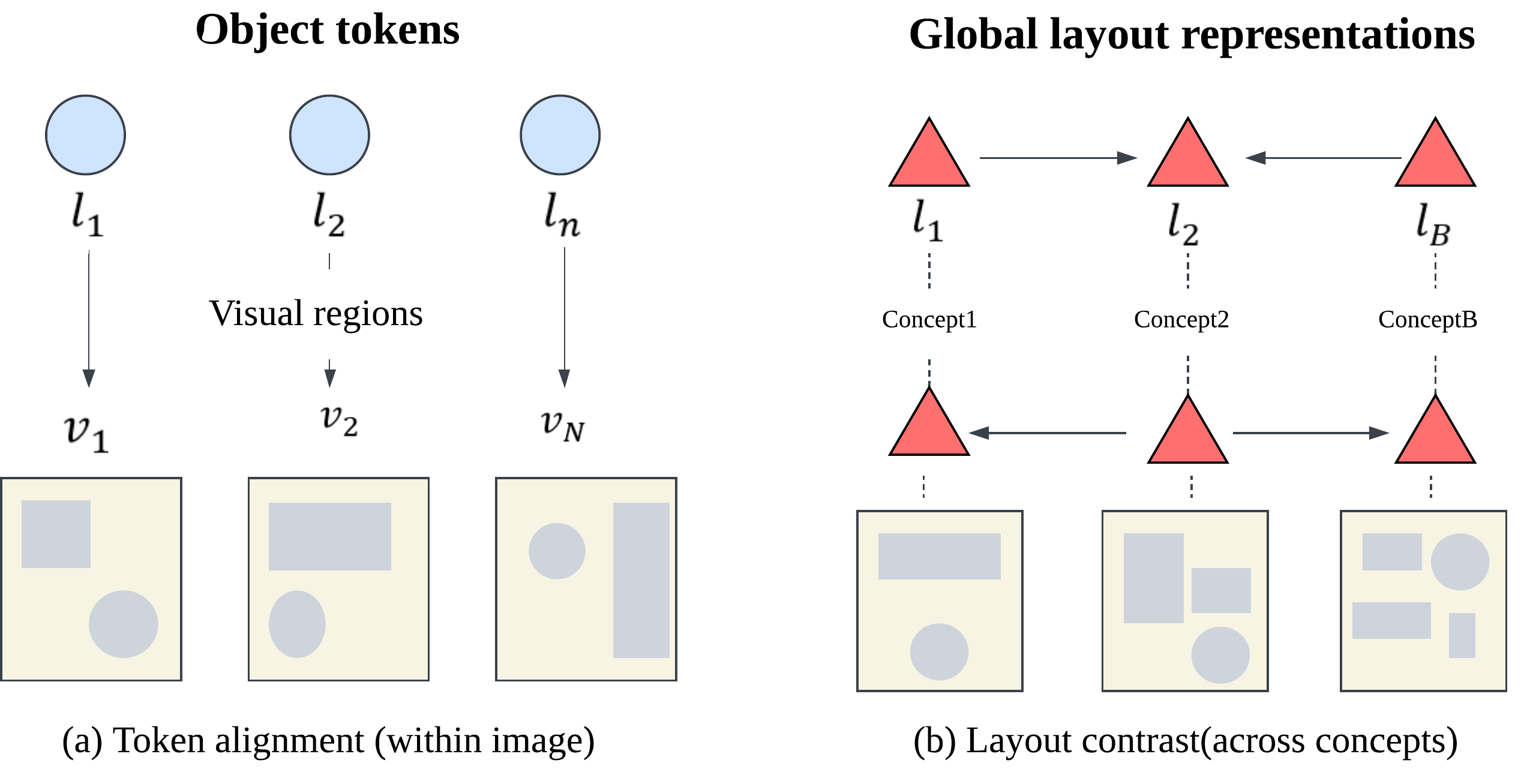}
\caption{Illustration of the two self-supervised learning objectives used in layout generation. (a) Token alignment: each predicted layout embedding $l_i$ is aligned with its corresponding visual region embedding $v_i$ extracted from real BookCover images. (b) Layout contrast: global embeddings $l_k$ are encouraged to be similar for augmented views of the same concept while being distinct across different STEM concepts.}
\label{fig2}
\end{figure}
In addition to this alignment, we introduce a complementary \textbf{layout contrastive loss} to enforce that semantically dissimilar STEM concepts yield distinct layout representations, while preserving intra-concept structural coherence. This promotes both conceptual clarity and pedagogical consistency. While the alignment loss operates within individual image–layout pairs, the layout contrastive loss operates across a batch of $B$ concept descriptions $\{c_k\}_{k=1}^B$. For each concept $c_k$, a global layout embedding $l_k = f_{\mathrm{layout}}(f_{\mathrm{text}}(c_k))$ is computed, and the contrastive loss is defined as:

\begin{equation}
\mathcal{L}_{\mathrm{laycontrast}} = -\frac{1}{N} \sum_{k=1}^{B} \log \frac{\exp(\mathrm{sim}(l_k, l_k^+)/\tau)}{\sum_{m=1}^{B} \exp(\mathrm{sim}(l_k, l_m)/\tau)}
\end{equation}

Here, $l_k^+$ denotes an augmented variant of $l_k$ (via stochastic masking or dropout), simulating alternate layout realizations of the same concept. In this formulation, the index $k$ identifies the anchor concept in the batch, while $m$ enumerates all concepts (including $k$ itself) as contrastive candidates in the denominator. This setup encourages layout embeddings of the same concept to cluster closely, while embeddings from different concepts remain separated, thereby enhancing both intra-class cohesion and inter-class discriminability in the learned layout space.

To further reinforce visual consistency across multiple instances of the same concept (e.g., in multi-frame narratives), we sample multiple layouts per concept $c_k$ and minimize their pairwise distance:

\begin{equation}
\mathcal{L}_{\mathrm{intra}} = \frac{1}{|P_k|^2} \sum_{i,j \in P_k} \left(1 - \mathrm{sim}(l_i, l_j)\right)
\end{equation}

where $P_k$ is the set of layout embeddings generated from different inputs or stochastic views of the same concept $c_k$. This intra-concept cohesion term acts as a soft structural anchor, promoting narrative stability and layout uniformity in educational sequences.

Our final layout contrastive objective becomes:

\begin{equation}
\mathcal{L}_{\mathrm{layout}} = \mathcal{L}_{\mathrm{laycontrast}} + \lambda_{\mathrm{intra}} \mathcal{L}_{\mathrm{intra}}
\end{equation}

As shown in Fig.~\ref{fig2}, part(a) illustrates token-level alignment within each layout-image pair, while part(b) depicts both inter-concept separation and intra-concept cohesion in the layout embedding space.

\subsection{Layout-to-Image Synthesis}

To transform structured layout representations into rich visual renderings, we adopt a diffusion-based image generator $G$ built upon a layout-conditioned variant of Stable Diffusion. As illustrated in Fig.~\ref{fig1} (top-center), the generator takes as input a global layout $L = \{(o_i, b_i)\}_{i=1}^N$ and a noise vector $z \sim \mathcal{N}(0,1)$, and produces a synthesized image $\hat{x}$ via the denoising process:
\begin{equation}
\hat{x} = G(L, z), \quad z \sim \mathcal{N}(0, I)
\end{equation}

Here, each $o_i$ is an object label and $b_i \in \mathbb{R}^4$ is the associated bounding box encoding the object's normalized spatial position and size. The layout $L$ is injected into the diffusion U-Net through cross-attention at multiple layers, enabling the generator to respect both the content and composition implied by the layout.

To enable layout-aware generation, each layout element—defined by an object label $o_i$ and its bounding box $b_i$—is encoded into a layout embedding $l_i \in \mathbb{R}^d$ using a combination of CLIP-based semantic encoding and positional embedding:
\begin{equation}
l_i = f_{{label}}(o_i) + f_{{pos}}(b_i)
\end{equation}

The resulting layout embeddings $l_i$ are injected into selected layers of the U-Net via masked cross-attention:
\begin{equation}
\mathcal{L}_{{inject}} = \{{mid-block}, {attn}_{{down}}, {attn}_{{up}} \}
\end{equation}

At each injection point, the U-Net queries $Q$ attend to layout embeddings $L = \{l_i\}_{i=1}^N$:
\begin{equation}
{Attn}(Q, L, M) = {Softmax}\left( \frac{Q L^\top + M}{\sqrt{d}} \right) L
\end{equation}

Here, $M$ is a spatial attention mask derived from $b_i$, assigning zero weights to positions within each bounding box and $-\infty$ elsewhere. This mask ensures that attention is spatially constrained, improving alignment between layout semantics and visual regions.

To ensure that the generated image $\hat{x}$ faithfully conveys the semantic essence of the original concept prompt $c$, we introduce a CLIP-based semantic alignment loss that compares the text embedding $f_{text}(c)$ with the visual embedding $f_{image}(\hat{x})$:
\begin{equation}
\mathcal{L}_{align} = 1 - \cos(f_{text}(c), f_{image}(\hat{x}))
\end{equation}

This loss encourages semantic consistency between the input prompt and generated output, effectively closing the loop between concept understanding and visual realization. As shown in Fig.~\ref{fig1}, this loss is implemented by passing the generated image through a frozen CLIP image encoder and computing cosine similarity against the prompt representation. Combined with the layout alignment objectives, this step enforces both structural and conceptual fidelity in the visual output.

\subsection{Knowledge-Driven Iterative Visualization}

To support progressive STEM instruction, we introduce a \textbf{knowledge-driven traversal module} that decomposes complex concepts into scaffolded sub-concepts. These sub-concepts are sequentially rendered using our pipeline, forming an interpretable visual reasoning chain. As depicted in Fig.~\ref{fig1} (left-top), the system leverages structured knowledge learned from BookCover–caption pairs, which encode how abstract themes are visually composed and narrated.

Formally, let $G_{STEM} = (\mathcal{C}, \xi)$ denote a domain-specific concept graph, where $\mathcal{C}$ is the set of instructional concepts and $\xi \subset \mathcal{C} \times \mathcal{C}$ defines pedagogical or prerequisite relations. For any high-level concept node $c_0$, we recursively apply our LEARN framework to generate a sequence of layout–image pairs aligned with instructional goals:
\begin{equation}
\{(L_i, \hat{x}_i)\} = LEARN(G_{STEM}, c_0)
\end{equation}

This iterative process supports step-wise abstraction, allowing agents to generate explanatory frames for increasingly sophisticated concepts. The traversal strategy follows a curriculum-informed ordering, ensuring that visual content aligns with the learner's cognitive progression: starting from foundational understanding and gradually moving toward analysis, application, and creative synthesis, in line with Bloom's taxonomy.

The entire process is informed by the conceptual–visual correspondences encoded in the learned BookCover embedding space, allowing the system to reuse familiar composition patterns while adapting to domain-specific semantics. This integration ensures both cognitive efficiency and instructional relevance, supporting STEM educators in the automatic construction of multi-frame visual narratives.

\section{Experiments}

We evaluate the \textbf{LEARN} framework across multiple dimensions tailored to the needs of layout-aware educational visual generation. These include: (1) spatial and semantic consistency across conceptually linked scenes, (2) fidelity of visual elements relative to structured STEM prompts, and (3) pedagogical suitability in supporting stepwise conceptual narration. Our evaluation setup builds upon layout-to-image (L2I) benchmarks with modifications relevant to STEM education. In addition to adopting regional layout diffusion strategies~\cite{li2023gligen,zheng2023layoutdiffusion},we introduce structure-sensitive metrics and controlled prompt tuning mechanisms to assess instructional coherence and the clarity of object arrangement in generated visuals.

\subsection{Evaluation Objectives and Metrics}
Our evaluation is guided by the following questions:
\begin{itemize}
\item Does LEARN generate visual narratives where repeated elements (e.g., recurring objects or characters) are rendered consistently across educational sequences?
\item How effectively does LEARN preserve layout integrity and visual clarity under educational constraints such as white backgrounds and structured positioning?
\item Do the proposed layout alignment, contrastive structure loss, and prompt modulation techniques enhance interpretability and instructional progression?
\end{itemize}

We report core metrics that best capture the structural and semantic precision essential to educational image generation:
\begin{itemize}
\item \textbf{Fr\'echet Inception Distance (FID)} $\downarrow$: Measures overall image coherence and realism. Although widely used, FID does not directly reflect pedagogical utility, which relies more on structural clarity and conceptual alignment.
\item \textbf{CropCLIP Score} $\uparrow$: Evaluates region-level semantic alignment between image subpatches and prompt concepts.
\item \textbf{SAMIoU} $\uparrow$: Measures alignment between predicted semantic regions and reference masks, reflecting structural accuracy.
\end{itemize}

These metrics are computed on the RC-COCO dataset after fine-tuning all compared models on the curated BookCover corpus. This setup enables a fair assessment of structure-grounded generation performance in concept-dense, pedagogically aligned image contexts.

\subsection{Implementation Details}

LEARN uses a CLIP ViT-B/32 text encoder (output dim 512) and a transformer-based layout decoder to predict up to 40 layout embeddings. Each token is projected to 768 dimensions to match CLIP region embeddings. Layout features are fused into a diffusion U-Net at 64$\times$64, 32$\times$32, and 16$\times$16 scales via GLIGEN-style masked cross-attention~\cite{li2023gligen,zheng2023layoutdiffusion}.

We train using AdamW (lr=1e-4, batch size 32) with early stopping on validation CLIPScore. Loss weights are: $\lambda_{\mathrm{align}}{=}1.0$, $\lambda_{\mathrm{laycontrast}}{=}0.5$, $\lambda_{\mathrm{semantic}}{=}1.0$, and $\lambda_{\mathrm{intra}}{=}0.36$.

To enhance consistency across related layouts (e.g., multi-frame scenes), we incorporate a lightweight variant of positive-negative prompt tuning (PNPT)~\cite{dong2022dreamartist}. Positive embeddings reinforce shared traits, while negative counterparts suppress irrelevant variance, improving detail fidelity without sacrificing diversity.

For diagrams requiring clear visual structure, we apply a soft background constraint by training pseudo-prompts initialized from empty descriptions. Their embeddings are refined using visual clarity metrics (e.g., luminance variance, edge clutter index), ensuring high-contrast and low-noise outputs. These prompt-level modulations, implemented via adapter tuning rather than full reinforcement learning, support controllable layout rendering while preserving narrative continuity and style alignment.

\subsection{Layout Fidelity and Narrative Support: Quantitative Insights}

Most layout–to–image (L2I) work pursues photorealism or generic caption fidelity, whereas LEARN targets \emph{instructional} fidelity: every object must sit in the right place, frame-to-frame, so that a learner can follow the physics or biology being illustrated.  We therefore report two small yet task-critical studies on the Rich-Context COCO (RC-COCO) benchmark.  All models, including the recent rich-context SDXL variant of \cite{cheng2024rethinking}, were \emph{first} fine-tuned on our BookCover+textbooks(30K+ images or so) corpus to equalize exposure to pedagogical layouts, then evaluated on RC-COCO.  

Table~\ref{tab:ablation} shows that removing layout embeddings or either alignment loss significantly degrades both region accuracy (SAMIoU) and text–patch alignment (CropCLIP), confirming their importance for producing precise and interpretable STEM visuals. Dropping BookCover fine-tuning also reduces structural fidelity, highlighting the value of curriculum-aligned visual priors for pedagogically coherent generation.

Table~\ref{tab:baselines} compares LEARN with two layout-to-image baselines fine-tuned on the BookCover dataset. LEARN achieves the best FID, indicating superior image realism, while maintaining competitive layout accuracy in CropCLIP and SAMIoU. Although SDXL slightly leads in layout metrics, LEARN offers better overall visual coherence without sacrificing spatial alignment. These qualities result in scenes where, for example, weights appear on the correct lever arm or magnets align with rails—details essential for STEM understanding yet underrepresented by small metric differences.

\begin{table}[t]
\caption{Ablation study on RC-COCO.  Removing any structural loss or token degrades both semantic and spatial quality.}
\label{tab:ablation}
\centering
\begin{tabular}{|l|c|c|c|}
\hline
Method & FID $\downarrow$ & CropCLIP $\uparrow$ & SAMIoU $\uparrow$ \\ \hline
\textbf{LEARN (full)}           & \textbf{27.16} & \textbf{27.92} & \textbf{81.52} \\
w/o layout embeddings               & 36.45          & 24.63          & 76.36\\
w/o $\mathcal{L}_{{align}}$      & 29.17          & 26.14          & 78.30\\
w/o $\mathcal{L}_{{laycontrast}}$ & 28.62          & 26.73          & 77.21\\ 
w/o BookCover fine-tuning & 29.52          & 26.31          & 80.37\\
\hline
\end{tabular}
\end{table}

\begin{table}[t]
\caption{Comparison with rich-context L2I baselines (all fine-tuned on book-covers).  LEARN pairs the best realism (FID) with competitive layout accuracy.}
\label{tab:baselines}
\centering
\begin{tabular}{|l|c|c|c|}
\hline
Model & FID $\downarrow$ & CropCLIP $\uparrow$ & SAMIoU $\uparrow$ \\ \hline
GLIGEN~\cite{li2023gligen}                    & 27.95 & 26.13 & 77.60 \\
SDXL (Rich-Context)~\cite{cheng2024rethinking} & 27.41 & \textbf{28.15} & 80.87 \\
\textbf{LEARN (ours)}                        & \textbf{27.16} & 27.92 & \textbf{81.52} \\ \hline
\end{tabular}
\end{table}

\subsection{Concept Progression and Structural Consistency}

This section shows qualitative and structural results on three STEM prompts: ``lever principle,'' ``cyclotron accelerator states,'' and ``bar magnet on inclined plane''. Each is illustrated through layout-conditioned sequences that highlight spatial causality and concept progression.
\begin{figure}[t]
\centering
\includegraphics[width=\textwidth, trim=0cm 0.1cm 0cm 0cm, clip]{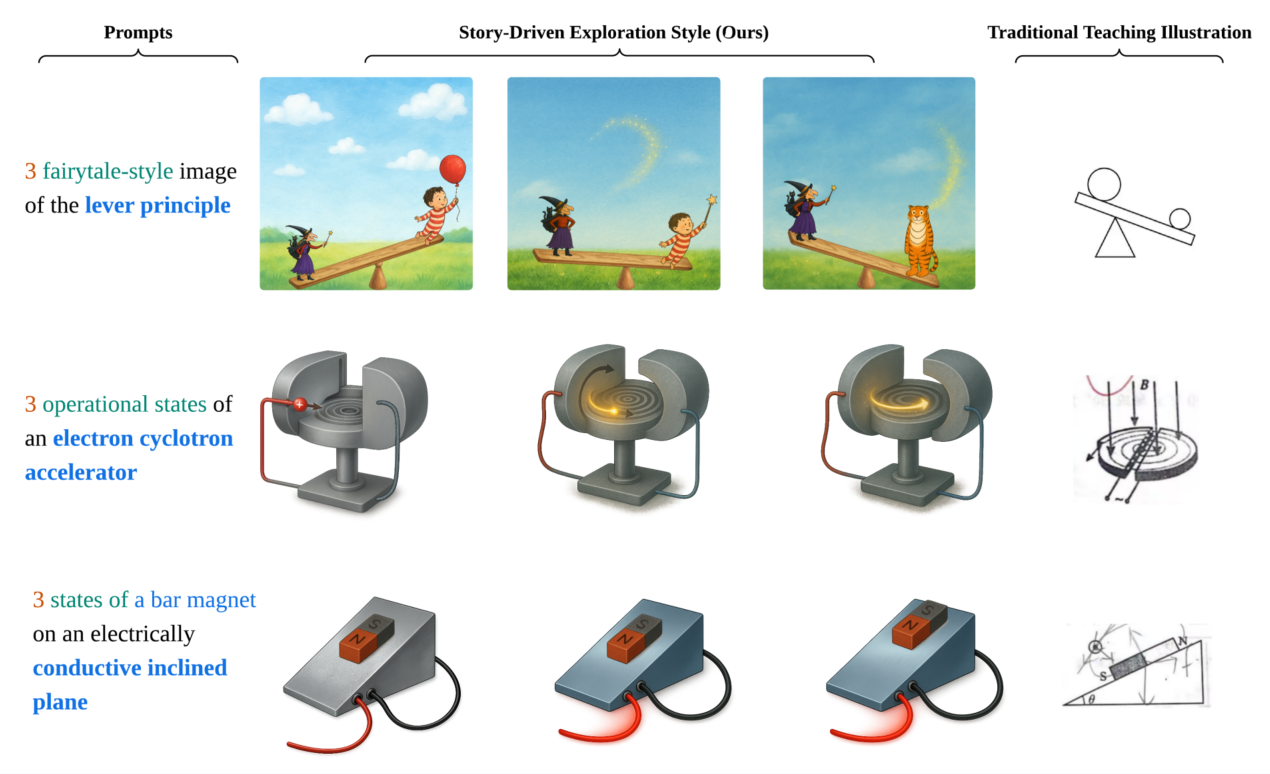}
\caption{LEARN-generated visual narratives for STEM prompts: (a) lever principle, (b) cyclotron accelerator states, and (c) magnet on inclined plane. Each sequence illustrates concept progression and spatial causality, supporting instructional clarity and exploration.}
\label{fig3}
\end{figure}

These visuals are not model-explained but demonstrate how teachers might frame concepts through structured storytelling. The model generates semantically rich visual material that can inspire classroom dialogue, scaffold reasoning, or serve as components in interactive digital tools. 
\begin{figure}[t]
\centering
\includegraphics[width=\linewidth]{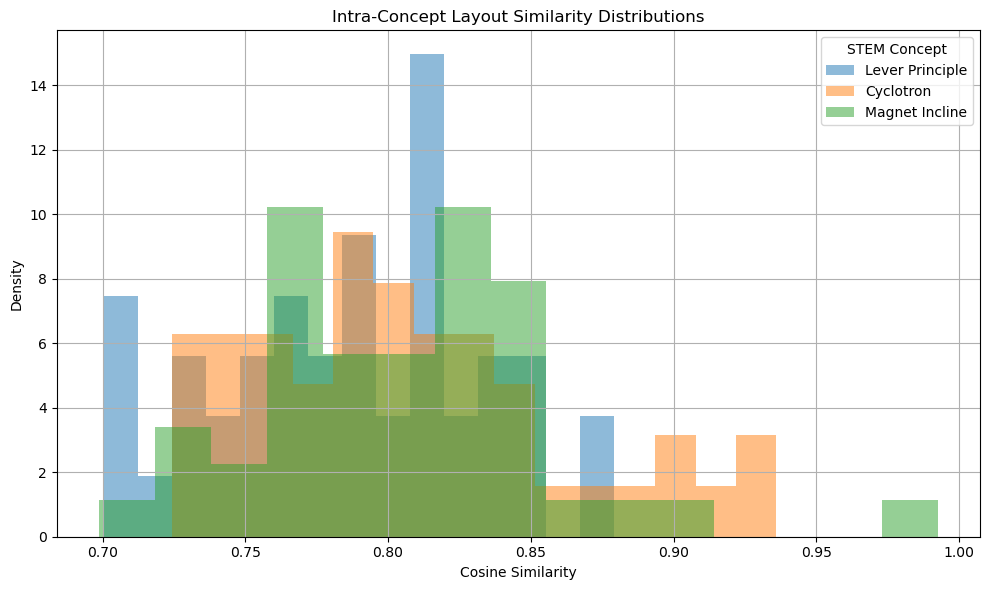}
\caption{Distribution of pairwise cosine similarities among layout embeddings generated for the same concept. Higher intra-concept similarity reflects stronger structural consistency across samples.}
\label{fig4}
\end{figure}

In Fig.~\ref{fig3}(a), a child and a witch on a seesaw illustrate shifting balances: from balloon-induced lift to wand intervention to a tiger transformation—each prompting questions on torque, mass, and force. Fig.~\ref{fig3}(b) captures how electrons spiral in a cyclotron, reinforcing magnetic field mechanics. Fig.~\ref{fig3}(c) traces a bar magnet’s motion along a conductive slope, prompting reasoning about friction and induction.

To quantify the structural coherence illustrated above, we measure pairwise cosine similarity between layout embeddings generated for the same concept. Fig.~\ref{fig4} shows the similarity distribution across concepts. Higher intra-concept similarity indicates more consistent layout structures. This aligns with LEARN’s narrative goal of producing visually coherent and pedagogically stable representations.
\section{Educational Evaluation}

To assess LEARN’s educational utility, we conducted: (1) a qualitative analysis of visualized STEM prompts and (2) a user study involving teachers and students.

\subsection{Qualitative Analysis}

As illustrated in Figure~\ref{fig3}, LEARN generates layout-guided visual sequences for prompts such as \textit{lever principle}, \textit{cyclotron accelerator states}, and \textit{bar magnet on inclined plane}. These outputs scaffold conceptual understanding by maintaining spatial consistency, representing physical causality, and encouraging exploratory thinking. While the model produces visuals only, teachers can use them to prompt layered inquiry (e.g., torque, induction) or embed them into interactive tools.

To summarize LEARN’s alignment with cognitive and instructional theories, Table~\ref{tab:clt_bloom} maps its visual strategies to Cognitive Load Theory (CLT) and Bloom’s taxonomy. Specifically, layout embedding helps reduce spatial load and aids procedural understanding. Multi-frame storytelling supports progressive reasoning, while consistency via PNPT reduces perceptual effort and facilitates higher-order learning tasks like synthesis and transfer.

\begin{table}[ht]
\centering
\caption{LEARN’s visual scaffolds mapped to CLT and Bloom’s taxonomy.}
\label{tab:clt_bloom}
\begin{tabular}{|l|l|l|}
\hline
\textbf{Visual Strategy} & \textbf{CLT Support} & \textbf{Bloom’s Level} \\
\hline
Layout embedding alignment & Reduces spatial load & Understanding, Application \\
Sequential frames & Structures cognitive progression & Analysis, Reasoning \\
Consistent traits (PNPT) & Lowers recognition cost & Transfer, Synthesis \\
\hline
\end{tabular}
\end{table}

\subsection{Human Study}

A user study with 12 educators (8 female, 4 male) and 26 students (10 female, 16 male, ages 11--22) evaluated 50 STEM prompts rendered by LEARN, GLIGEN, and textbook figures. On a 5-point Likert scale~\cite{jamieson2004likert}, LEARN’s outputs scored 23\% higher in clarity and reduced perceived cognitive load by 31\%. Moreover, 96\% of the participants reported stronger narrative flow and improved concept alignment.

Educators highlighted enhanced clarity in explaining abstract content (e.g., field lines, polarity), while students noted better visual consistency and interpretability. Taken together, these findings suggest that LEARN supports instructional coherence, facilitates layered reasoning, and promotes cognitively aligned learning pathways.

\section{Limitations and Future Work}

While LEARN significantly enhances the narrative structure and layout accuracy of generated STEM visuals, several limitations remain. First, not all concepts lend themselves to clean segmentation into meaningful sub-scenes. Some visual outputs, despite structural coherence, fail to express stepwise pedagogical logic. Enriching concept descriptions with large language models (LLMs) like ChatGPT-4o helps alleviate this issue. However, how to effectively align such LLM-generated semantics with BookCover layouts, curricular goals, and individual learner needs remains a key challenge for scalable and personalized deployment.

Second, broader deployment in real-world educational settings still poses challenges. One obstacle is bridging the gap between generative outputs and classroom usability—especially how students and teachers can easily integrate such tools into open-ended exploration or self-paced learning environments. Another challenge is advancing beyond static storyboards toward more dynamic, simulation-based representations that can capture fine-grained scientific phenomena. Additionally, integrating LEARN more tightly with increasingly capable multimodal LLMs presents a promising but open frontier. Doing so could unlock deeper forms of personalized reasoning and enable one-stop, multi-modal inquiry pathways for learners.

Addressing these challenges will require advances not only in model design but also in human-AI interaction paradigms, curriculum alignment, and educational deployment strategies.
\section{Conclusion}
This study introduces LEARN, a layout-aware diffusion framework that integrates narrative structure into visual generation to meet emerging educational needs in the GenAI era. The framework is supported by a curated and enriched BookCover dataset, enabling the model to learn implicit relationships between visual composition and conceptual meaning. Evaluations on layout accuracy and region-level alignment demonstrate that LEARN maintains structural fidelity, while user studies with teachers and students show improved interpretability, reduced cognitive load, and increased engagement. These findings highlight LEARN’s potential as a practical tool for STEM instruction and as a bridge between generative AI, cognitive theory, and curriculum-based learning. Future work may explore linking LEARN with LLM-driven curriculum parsers or adaptive tutoring systems to enable interactive, query-driven visual exploration and dynamic scene generation for personalized education.

\begin{credits}
\subsubsection{\ackname} We thank Zhang Wei, currently a teacher, from No.1 Senior Middle School of Shouguang for organizing students and teachers participation in our model evaluation sessions. His feedback from a classroom teaching perspective was invaluable in shaping the educational direction of this work. This research was supported in part by a JSPS KAKENHI Grant Number JP23K11170.

\subsubsection{\discintname}
The authors have no competing interests to declare that are
relevant to the content of this article.
\end{credits}
\bibliographystyle{splncs04}
\bibliography{mycite}

\end{document}